# Evaluating ChatGPT's Performance for Multilingual and Emoji-based Hate Speech Detection

**Mithun Das, Saurabh Kumar Pandey, Animesh Mukherjee**
Indian Institute of Technology, Kharagpur, India
mithundas@iitkgp.ac.in, saurabh2000.iitkgp@gmail.com, animeshm@cse.iitkgp.ac.in

## Abstract

Hate speech is a severe issue that affects many online platforms. So far, several studies have been performed to develop robust hate speech detection systems. Large language models like ChatGPT have recently shown a great promise in performing several tasks, including hate speech detection. However, it is crucial to comprehend the limitations of these models to build robust hate speech detection systems. To bridge this gap, our study aims to evaluate the strengths and weaknesses of the ChatGPT model in detecting hate speech at a granular level across 11 languages. Our evaluation employs a series of *functionality tests* that reveals various intricate failures of the model which the aggregate metrics like macro F1 or accuracy are not able to unfold. In addition, we investigate the influence of complex emotions, such as the use of emojis in hate speech, on the performance of the ChatGPT model. Our analysis highlights the shortcomings of the generative models in detecting certain types of hate speech and highlighting the need for further research and improvements in the workings of these models.

## 1 Introduction

Hate speech is a significant problem on social media platforms, which can cause emotional harm, social division, and even violence (Röttger et al., 2022; Das et al., 2022a). Therefore, the development of effective hate speech detection systems is crucial to mitigate the negative impacts of hate speech. So far, several works have been done to develop hate speech detection models, and several datasets have been proposed in multiple languages to build robust detection systems(Fortuna and Nunes, 2018; MacAvaney et al., 2019; Parikh et al., 2021). The way these models are evaluated generally involves keeping a held-out or test dataset separate from the created data, and then the model's performance is checked on the test data (Waseem and Hovy, 2016; Banerjee et al., 2021). However, the problem with this technique is that if the test set does not have a sufficient representation of diverse hate speech, the model may exhibit good performance, which is not representative of the true case as the model is not being evaluated in a holistic fashion. The reason for the lack of diversity in the test set can be attributed to the way they are sampled (Röttger et al., 2021). Generally, these datasets are created by scraping social media posts based on certain hateful lexicons or target community names, which may lead to the potential miss of diverse types of hate speech (Das et al., 2022c).

Therefore to find out the limitation of the existing models, researchers have introduced novel test sets and methods that allow for a more sophisticated evaluation of model functionalities (Ribeiro et al., 2020; Röttger et al., 2022). Model functionalities refer to the specific tasks or functions that a machine learning model is designed to perform (Röttger et al., 2021; Das et al., 2022c; Kirk et al., 2022). These tasks can vary depending on the application and the type of model being used. Evaluating the functionalities of a model is important in determining the performance and effectiveness of the model. It helps in identifying areas for improvement, optimizing model parameters, and developing more accurate and robust models.

Recently, pre-trained language models, such as ChatGPT (OpenAI, 2023a), have shown great potential in performing several tasks, including hate speech detection (Zhu et al., 2023; Huang et al., 2023). It has been demonstrated that ChatGPT can achieve an accuracy of approximately 80% when compared to MTurker annotations (Li et al., 2023). While language models like ChatGPT have shown promising results in detecting hate speech, there is a need to investigate their limitations to ensure that these models are reliable and robust. Furthermore, it remains unclear how this model will perform for low-resource languages. In addition, hate speech detection can be tricky when complex emotions, actions, and intentions are conveyed in the post. One example of such emotions is the use of emojis in the post. Studies have shown that over 95% of Internet users use emojis, and 10 million emojis are sent daily (Brandwatch, 2018). Therefore, we aim to explore the limitations of the ChatGPT model by answering the following two questions.

**RQ1** How effective is ChatGPT based on a diverse set of *functionality tests* in detecting hate speech across languages?

**RQ2** What are the weaknesses of ChatGPT in detecting emoji-based hate speech?

To answer **RQ1**, we utilize the Multilingual HateCheck (MHC) (Röttger et al., 2022) framework, which consists of a suite of functional tests designed to evaluate the robustness of the low-resource hate speech detection model in ten languages. The authors created several test cases that map to various functionalities to understand the weaknesses present in a model. We also incorporate the original HateCheck functionalities, originally developed for English (Röttger et al., 2021).

To answer **RQ2**, we used the HATEMOJICHECK (Kirk et al., 2022), which is designed to evaluate the emoji-based hate speech detection model. The authors provided 3,930 test cases for seven functionalities covering six identities to explore critical model weaknesses. We passed these test cases through the ChatGPT model, recorded the predictions, and calculated the accuracy achieved for different functionalities.

**Key Observations:** While the ChatGPT model's performance is excellent for detecting hateful posts, it fails to identify non-hateful counterspeeches and often misclassifies them as hate speech. Additionally, the model's ability to distinguish between protected and non-protected target groups is more effective in English compared to non-English languages, leading to the misclassification of abusive content towards non-protected groups as hate speech for non-English languages. In the case of Emoji-based hate speech, the model exhibits inadequate performance when positive emojis are used in a hateful post, which poses challenges in accurately determining the appropriate label for such instances.

## 2 Related Work

In this section, we will review some of the existing studies on hate speech detection and its evaluation, as well as the research conducted around the ChatGPT model.

**Hate Speech Detection:** A significant amount of work has been done to develop hate speech detection models for multiple languages. While the majority of datasets are in English (Waseem and Hovy, 2016; Davidson et al., 2017; de Gibert et al., 2018; Kumar et al., 2018), several multilingual datasets (Pitenis et al., 2020) have emerged in the research community to build hate speech detection systems. In fact, several shared task competitions have been conducted to encourage research in non-English languages. For instance, Mandl et al. (2019) organized several shared tasks to detect hate speech in Indo-Aryan languages, while Mulki and Ghanem (2021) conducted the Arabic Misogyny Identification shared task. Additionally, various multilingual hate speech datasets have been made available for research in languages such as Arabic and French (Ousidhoum et al., 2019), Bengali (Das et al., 2022b), Hindi (Bohra et al., 2018), Spanish (Basile et al., 2019), Indonesian (Ibrohim and Budi, 2019), Italian (Sanguinetti et al., 2018), Polish (Ptaszynski et al., 2019) and Portuguese (Fortuna et al., 2019).

**Evaluating Hate Speech Detection Systems:** Although the concept of functional testing in software engineering has been around for a long time (Beizer, 1995), Ribeiro et al. (2020) introduced the idea of functional tests in natural language processing to evaluate models and show how their method can detect the strengths and weaknesses of models at a granular level that is often obscured by high-level metrics such as accuracy and F1-score. Building on prior research, Röttger et al. (2021) developed a set of test cases covering 29 functionalities to evaluate hate speech detection systems in English and named their test suite HATECHECK. Das et al. (2022c) adapted the framework of HATECHECK and built HateCheckHIn, introducing six new functionalities to evaluate Hindi hate speech detection systems. Kirk et al. (2022) also adapted the same framework and built HATEMOJICHECK to detect emoji-based hate speech. Furthermore, Röttger et al. (2022) introduced MHC, covering 34 functionalities in ten languages, which is more languages than any other hate speech evaluation dataset. Therefore, in our work, we use HATECHECK & MHC to evaluate ChatGPT's performance across languages and HATEMOJICHECK to evaluate ChatGPT's per-formance on emoji-based hate speech.

**ChatGPT:** ChatGPT (OpenAI, 2023a) is a state-of-the-art large language model developed by OpenAI, based on the GPT-3.5 architecture. It is capable of comprehending and generating text that resembles human speech in various languages and domains. ChatGPT has a wide range of natural language processing abilities, including text generation (Chen et al., 2023), question-answering (Omar et al., 2023), language translation (Jiao et al., 2023), and summarization (Yang et al., 2023). The model can learn and generalize linguistic patterns through its training on a vast amount of data. ChatGPT has demonstrated excellent performance in various language tasks, including hate speech identification (Zhu et al., 2023; Huang et al., 2023; Li et al., 2023). Therefore, this study aims to evaluate ChatGPT's performance in detecting hate speech at a granular level.

## 3 Functional Testing

### 3.1 Defining Functional Tests

In the context of hate speech, functional testing refers to the ability of a hate speech detection model to classify hateful and non-hateful posts of different types correctly (Röttger et al., 2021). Each functionality is defined by a set of test cases sharing a common gold-standard label. For instance, when the sentence "I hate women" is passed to a hate speech detection model, the model should correctly classify the post as hateful. Conversely, a sentence like "I hate pizza" should be classified as non-hateful. Functional testing aims to evaluate the hate speech detection model's performance at a granular level, testing its ability to identify specific types of hateful content and distinguish them from non-hateful content.

### 3.2 (MULTILINGUAL) HATECHECK

The MHC test suite (Röttger et al., 2022) comprises a total of 34 functionalities, expanded from work conducted by Röttger et al. (2021), covering ten languages with a total of 36,582 test cases. These functionalities were selected based on interviews with civil society stakeholders and a review of hate speech research. Native-speaking language experts were hired to handcraft the test cases for these functionalities. To provide a better understanding, let's summarize the functionalities. The **F1-F4** functionalities check how the model classifies derogatory and hateful posts. The **F5-F6** functionalities evaluate the model's performance for threatening language. **F7** checks how the model performs on hateful posts made using slurs. **F8-F9** evaluates the model's performance on using profanity in both hateful and non-hateful contexts. **F10-F11** evaluates how the model performs on hateful posts expressed through reference in subsequent clauses and sentences. **F12-F13** evaluates the model's performance using negated expressions in hateful and non-hateful contexts. **F14-F15** evaluates how the model performs on using hateful posts phrased as a question or opinion. **F16-F17** evaluates the usage of protected group identifiers in neutral and positive statements. **F18-F19** evaluates the usage of announcements of hateful posts through counter-speech. **F20-F22** evaluates the model's performance in the use of abuse against non-protected targets. **F23-F24** evaluates how the model performs on hateful posts with spelling variations such as swapping adjacent characters, missing characters, missing word boundaries, Leet speak spellings, etc (see Appendix A for dataset statistics)[1].

### 3.3 HATEMOJICHECK

HATEMOJICHECK (Kirk et al., 2022) is a test suite designed for functional testing of emoji-based hate speech detection, with a total of 7 functionalities in English. The author developed these functionalities based on existing research to capture real-world uses of emoji-based hate speech, covering distinct aspects of unambiguous hate speech. **F1 Verb Swap,** tests the model's performance when verbs are swapped with their equivalent emojis (👊, 💣, 🔪, 🪓). In **F2 Identity Swap**, representative emojis are used instead of identity names(Woman:👩, Black People:👨🏿, Gay People:🏳️‍🌈) in hateful posts. **F3 Descriptor Swap** replaces nouns or emotions with matching emojis(💩, 🤮, 😷, 🦍, 🐪, 🐀). In **F4 Double Swap**, **F1** is combined with **F2** or **F3**. **F5 Append** evaluates the insertion of negative emotion(😤, 😡, 😠, 🤬) with neutral text. **F6 Positive Confounder** examines the use of positive emojis(😊, 😇, 😍, ☺️) in hateful texts. Finally, **F7 Emoji Leetspeak** replaces characters or word pieces(x:❌, i:1️⃣, o:0️⃣) with emojis while retaining the text's meaning. To enhance the robustness of the functional test suites, the author incorporated three types of perturbations for each functionality: **identity perturbations** (substituting the targeted identity with a non-protected entity), **polarity perturbations** (reversing the negative sentiment of the original hateful statement to make it positive and non-hateful), and **no emoji perturbations** (removing or replacing the emojis with their equivalent text to preserve the semantic expression). The HATEMOJICHECK test suite comprises a total of 3,930 entries, of which 2,126 are original test cases, and 1,804 are perturbations.

---

[1] HateCheck includes two additional functionalities, ***F8:****Non-hateful homonyms of slurs* and ***F9:****Reclaimed slurs* for the English language, which were excluded from MHC test suite. We refer them as **F8*** & **F9*** respectively.

## 4 Testing ChatGPT Across Functionalities

### 4.1 The ChatGPT model

As the base model, we employ the "**gpt-3.5-turbo**" model (OpenAI, 2023b), a large language (LLM) chatbot based on the GPT-3.5 language model developed by OpenAI. It has been optimized especially for chat-based applications but performs competently for traditional completion tasks, and the same model is utilized in the ChatGPT product. This model was chosen for its outstanding performance and capability to generate high-quality text. It is considered the most efficient and cost-effective of all chat-based models. For consistency. To use the model, OpenAI provides several parameters, such as max_tokens, temperature, and top_p, which users can modify in the API request according to the task that needs to be performed. The temperature parameter (a value between 0 and 2) controls the level of randomness in the generated response. As we intend to perform a binary classification task, we set the temperature value to 0 to reduce the randomness, ensuring a more consistent output from the model. We use the default values for other controllable parameters. We refer to the "**gpt-3.5-turbo**" model as "ChatGPT" throughout the paper.

### 4.2 Prompt Designing

A prompt is a set of instructions provided to an LLM that customizes and enhances its capabilities for a specific task (White et al., 2023). To effectively communicate with Large Language Models such as ChatGPT, it is necessary to define clear prompts particular to the task. The need for designing good prompts is that they can significantly impact the quality and accuracy of the model's outputs (Li et al., 2023). A well-designed prompt can help the model produce more relevant and coherent responses, while a poorly designed prompt can lead to irrelevant or nonsensical outcomes. Besides, a good prompt can help mitigate bias and toxicity in the model's outputs by encouraging the model to generate more socially acceptable responses. Therefore, carefully crafting prompts is essential in using LLMs for various natural language processing tasks.

In our case, the purpose is to classify statements as hate speech or non-hate, so we have crafted the following prompt: *"Can you determine if the given statement would be classified as hate speech according to the provided definition? [Hate Speech Definition]. Answer only 'Yes' or 'No'. Statement: <input text>".*

**Hate Speech Definition:** Following prior research (Röttger et al., 2021, 2022; Kirk et al., 2022), hate speech is defined as abuse that targets a protected group or its members for being part of that group. Protected groups include those based on age, disability, gender identity, race, national or ethnic origin, religion, sex, or sexual orientation, which reflects the international legal consensus. Based on the definition, we query the ChatGPT model through the API to classify content as hate or non-hate.

### 4.3 Results

We evaluate the model from several perspectives– (a) Performance Across Labels, (b) Performance Across Multilingual Functional Tests, (c) Performance Across Emoji-based Functional Tests, (d) Performance Across Target Groups, (e) Cases Where ChatGPT could not assign any label. The languages are represented by ISO 639-1 codes, while the emoji hate speech data is denoted as EMOJI (EMO). We highlight the performance below random choice (<50%) in blue. We also illustrate the percentage of data points ChatGPT could not label in (parenthesis).

**Performance Across Labels:** Table 1 depicts the performance of the ChatGPT model across all the languages, including the emoji-based hate speech detection results. We observe that ChatGPT exhibits diverse performances across the investigated languages. As expected, English attained the highest macro F1 score of 89.2. Additionally, we observe the superior performance of the ChatGPT model in languages such as Portuguese (87.1), Dutch (85.1), Spanish (84.2), Italian (83.7), German(83.6), and Mandarin (82.7). On the contrary, Hindi (67.3) and Arabic (71.6) depicted comparatively lower performance in the classification task across all languages.

Furthermore, we notice that the F1 score achieved for the hateful class was higher compared to the non-hate class. Although the F1 scores for the hateful class are impressive for Arabic and Hindi, the performance in the non-hate class is considerably inferior. This explains low macro F1 scores in these languages. Specifically, the difference in F1 scores for Hindi was over 55%, whereas, for Arabic, it was over 40%.

In addition, we also studied the percentage of posts for which the model could not assign any label. Interestingly, we found that the model experienced challenges in assigning labels for certain posts. The model left approximately 3.5% of the total posts unlabeled for the Arabic language. Similarly, for Hindi, around 1.9% of the posts were unlabeled. On the other hand, these percentages were significantly lower for German, Portuguese, and Spanish, indicating better performance in label assignments for these languages.

When evaluating the emoji-based hate speech detection, the model achieved an overall macro F1 score of 82.6. Similar to the multilingual setting, we observe that the F1 score for the hate class is higher than the non-hate class's. This indicates that the model demonstrates improved accuracy in identifying instances of hate speech, which is essential. However, it also raises concerns regarding the non-hate class as the false positive rate increases.

| Language | F1(h) | F1(nh) | Mac. F1 |
|---|---|---|---|
| **English/EN** | 99.7 | 78.6 | 89.2 |
| **Arabic / AR** | 93.3 (2.8) | 49.9 (5.3) | 71.6 (3.5) |
| **Dutch / NL** | 98.9 (0.2) | 71.4 | 85.1 (0.1) |
| **French / FR** | 99.0 (0.2) | 65.4 (0.1) | 82.2 (0.2) |
| **German / DE** | 99.5 (0.0) | 67.8 (0.2) | 83.6 (0.1) |
| **Hindi / HI** | 96.3 (1.2) | 38.3 (3.6) | 67.3 (1.9) |
| **Italian / IT** | 98.2 (0.2) | 69.2 | 83.7 (0.1) |
| **Mandarin / ZH** | 97.7 (0.5) | 67.7 (0.5) | 82.7 (0.5) |
| **Polish / PL** | 95.7 (1.0) | 67.2 (1.1) | 81.5 (1.1) |
| **Portuguese / PT** | 98.5 | 75.8 | 87.1 |
| **Spanish / ES** | 99.2 | 69.3 (0.2) | 84.2 (0.1) |
| **EMOJI/ EMO** | 88.6 | 76.6 (0.1) | 82.6 (0.1) |

Table 1: Performance across all the languages in terms of F1 score. h: hateful, nh: non-hateful.

**Performance Across Multilingual Functional Tests**: We present the performance of the multilingual functionalities in Table 2. We evaluate the ChatGPT model across various functional tests and identify specific weaknesses across the investigated languages. ChatGPT outperforms the random binary choice baseline (50% accuracy) on all functional tests for the hateful class. Especially in most languages, ChatGPT achieves a performance exceeding 90%, emphasizing its superior ability to detect and identify hateful content.

Additionally, we observe that the ChatGPT model demonstrates inferior performance for the counter-speech-related functionalities, suffering to distinguish between hate speech and counterspeech. The model exhibits below 50% performance for counterspeech-related functionality for almost all languages. Particularly for **F19**, the model attains 4.1% accuracy for the Hindi language, and for **F18**, it achieves a 1.4% accuracy for the Arabic language.

Further, we witness that for **F21**(abuse targeted at individuals not as a member of a protected group) functionality the model performance is quite less across languages, suggesting that abusive speech towards individuals is being misclassified, indicating a challenge for the model in distinguishing non-hateful abusive speech targeting individuals. Furthermore, we find that ChatGPT's performance for **F22**(abuse targeted at non-protected groups) is lower than a random binary choice baseline (50% accuracy) in almost all the languages except for English, French, Italian, and Portuguese. Even for French, Italian, and Portuguese, the performance is close to 50%, which signifies that while ChatGPT can distinguish non-protected groups in English, it struggles to do so for non-English languages, thus explaining instances where abusive content in a post is incorrectly labeled as hateful. Among all the languages, the only non-hateful functional test surpassing a random choice baseline is identifying abuse targeted towards objects(**F20**).

Furthermore, we observed cases where the model was unable to assign any label for specific functionalities. Specifically, for **F7**(hate expressed using slurs), **F9**(non-hateful use of profanity), and **F21**(abuse targeted at individuals, not as a member of a protected group), the model encountered challenges in labeling the content, albeit with variations across languages. For instance, for the **F9** functionality, the model could not assign any label for the Arabic language in 19% of the cases. Similarly, for **F7**, the model experienced difficulty labeling 10% of the samples for the Hindi language, highlighting the need for additional training, par-

| | Functionality | GL | Accuracy(%) | | | | | | | | | | |
|---|---|---|---|---|---|---|---|---|---|---|---|---|---|
| | | | EN | AR | NL | FR | DE | HI | IT | ZH | PL | PT | ES |
| Derogation | **F1:** Expression of strong negative emotions (explicit) | h | 99.3 | 100 | 99.3 | 100 | 99.3 | 95.7 | 96.4 | 100 | 97.9 | 100 | 100 |
| | **F2:** Description using very negative attributes (explicit) | h | 100 | 95.7 | 100 | 100 | 100 | 95.7 (2.9) | 100 | 100 | 100 | 100 | 100 |
| | **F3:** Dehumanisation (explicit) | h | 100 | 97.1 (1.4) | 100 | 100 | 100 | 100 | 100 | 100 | 100 | 100 | 100 |
| | **F4:** Implicit derogation | h | 97.1 | 89.0 (0.7) | 97.9 | 95.7 | 97.2 | 98.6 | 97.9 | 93.6 | 93.6 | 94.2 | 97.1 |
| Threat language | **F5:** Direct threat | h | 100 | 95.0 (0.7) | 100 | 100 | 99.3 | 100 | 100 | 100 | 100 | 100 | 100 |
| | **F6:** Threat as normative statement | h | 100 | 99.3 | 100 | 100 | 100 | 98.6 (1.4) | 99.3 | 100 | 100 | 100 | 100 |
| Slurs | **F7:** Hate expressed using slur | h | 99.3 | 82.4 (11.0) | 92.4 (1.8) | 100 | 99.2 | 82.2 (10.0) | 95.0 (0.7) | 86.7 (4.7) | 84.2 (5.3) | 91.4 | 98.0 |
| | **F8*:** Non-hateful homonyms of slurs | nh | 73.3 | - | - | - | - | - | - | - | - | - | - |
| | **F9*:** Reclaimed slurs | nh | 75.3 | - | - | - | - | - | - | - | - | - | - |
| Profanity usage | **F8:** Hate expressed using profanity | h | 100 | 100 | 100 | 99.2 (0.8) | 100 | 100 | 100 | 100 | 98.6 | 100 | 100 |
| | **F9:** Non-hateful use of profanity | nh | 98.0 | 62.0 (19.0) | 95.0 | 94.0 | 95.0 (2.0) | 53.0 (9.0) | 89.0 | 93.0 (2.0) | 85.0 (9.0) | 98.0 | 97.0 (1.0) |
| Pronoun reference | **F10:** Hate expressed through reference in subsequent clauses | h | 100 | 100 | 100 | 100 | 100 | 97.9 | 100 | 100 | 100 | 100 | 100 |
| | **F11:** Hate expressed through reference in subsequent sentences | h | 100 | 97.1 | 100 | 100 | 100 | 97.9 | 100 | 97.9 | 100 | 99.3 | 100 |
| Negation | **F12:** Hate expressed using negated positive statement | h | 100 | 92.9 | 99.3 | 100 | 97.9 | 95.7 | 100 | 100 | 100 | 100 | 100 |
| | **F13:** Non-hate expressed using negated hateful statement | nh | 91.0 | 40.7 (1.4) | 95.0 | 85.7 | 92.1 | 33.6 (1.4) | 95.7 | 85.7 | 85.0 (0.7) | 100 | 84.3 |
| Phrasing | **F14:** Hate phrased as a question | h | 100 | 93.6 | 100 | 99.3 (0.7) | 100 | 100 | 100 | 100 | 100 | 100 | 100 |
| | **F15:** Hate phrased as an opinion | h | 100 | 98.6 | 100 | 100 | 100 | 100 | 100 | 100 | 100 | 100 | 100 |
| Non-hateful group identifier | **F16:** Neutral statements using protected group identifiers | nh | 95.2 | 87.9 (0.7) | 92.9 | 83.6 | 80.7 | 50.7 (4.3) | 91.4 | 95.7 | 95.7 | 90.2 | 75.0 |
| | **F17:** Positive statements using protected group identifiers | nh | 100 | 78.1 (2.4) | 99.0 | 93.3 | 94.3 | 60.5 (1.0) | 93.8 | 92.9 | 93.8 | 98.1 | 97.1 |
| Counter speech | **F18:** Denouncements of hate that quote it | nh | 41.0 | 1.4 | 29.4 | 17.4 | 20.6 | 8.2 | 20.5 | 26.2 | 24.4 | 28.0 | 31.1 |
| | **F19:** Denouncements of hate that make direct reference to it | nh | 59.6 | 13.0 0.7 | 35.3 | 25.7 | 33.5 | 4.1 (0.7) | 31.1 | 28.0 (1.8) | 34.7 | 53.4 | 46.3 |
| Abuse against non-protected targets | **F20:** Abuse targeted at objects | nh | 100 | 83.1 (7.7) | 96.9 | 93.8 (1.5) | 96.9 | 80.0 (6.2) | 96.9 | 96.9 | 92.3 | 98.5 | 95.4 (1.5) |
| | **F21:** Abuse targeted at individuals (not as member of a protected group) | nh | 58.5 | 37.5 (28.1) | 53.8 | 60.0 | 46.2 | 32.3 13.8 | 58.5 | 44.6 (1.5) | 50.8 (4.6) | 56.9 | 44.6 |
| | **F22:** Abuse targeted at non-protected groups (e.g. professions) | nh | 75.8 | 49.2 (9.2) | 44.6 | 50.8 | 46.2 | 35.4 (9.2) | 52.3 | 46.2 | 49.2 | 55.4 | 44.6 |
| Spelling variations | **F23:** Swaps of adjacent characters | h | 100 | - | 100 | 99.3 | 100 | 99.3 | 97.1 | - | 97.1 | 98.6 | 97.9 |
| | **F24:** Missing characters | h | 100 | - | 95.0 | 97.9 | 100 | 86.4 (2.9) | 97.1 | - | 94.3 (0.7) | 97.1 | 96.4 |
| | **F25:** Missing word boundaries | h | 99.3 | - | 98.2 (0.6) | 94.0 (0.6) | 99.4 | 91.8 (2.1) | 93.2 (1.9) | - | 83.5 (5.1) | 96.3 | 96.3 |
| | **F26:** Added spaces between chars | h | 100 | 85.6 (6.8) | 100 | 100 | 100 | 96.6 (2.7) | 96.9 (0.6) | - | 90.9 (2.3) | 98.1 | 100 |
| | **F27:** Leet speak spellings | h | 100 | - | 99.4 | 97.0 (1.8) | 98.7 (0.6) | 92.5 (2.1) | 95.7 | - | 92.6 (2.3) | 97.5 | 100 |
| | **F28:** AR: Latin char. replacement | h | - | 85.7 (6.8) | - | - | - | - | - | - | - | - | - |
| | **F29:** AR: Repeated characters | h | - | 92.5 (3.4) | - | - | - | - | - | - | - | - | - |
| | **F30:** AR: Arabizi (Arabic chat alphabet) | h | - | 75.9 (20.3) | - | - | - | - | - | - | - | - | - |
| | **F31:** AR: Accepted alt. spellings | h | - | 95.9 (2.1) | - | - | - | - | - | - | - | - | - |
| | **F32:** ZH: Homophone char. replacement | h | - | - | - | - | - | - | - | 97.1 | - | - | - |
| | **F33:** ZH: Character decomposition | h | - | - | - | - | - | - | - | 93.8 (1.4) | - | - | - |
| | **F34:** ZH: Pinyin spelling | h | - | - | - | - | - | - | - | 94.6 (2.0) | - | - | - |

Table 2: Accuracy(%) across functional tests for each language. GL: Gold Label. h: hateful, nh: non-hateful.

| | Functionality | Label | *n* | Example Test Case Template | Accuracy(%) |
|---|---|---|---|---|---|
| ***F1*** | **Verb Swap** | h | 300 | I want to 🔪 all [IDENTITY] | 98.3 |
| ***F1.1*** | Identity perturbation | nh | 50 | I want to 🔪 all bigots | 40.0 |
| ***F1.2*** | Polarity perturbation | nh | 60 | I want to 🙏 all [IDENTITY] | 70.0 |
| ***F1.3*** | No emoji perturbation | h | 60 | I want to kill all [IDENTITY] | 85.0 |
| ***F2*** | **Identity Swap** | h | 120 | I hate [IDENTITY EMOJI] | 97.5 |
| ***F2.1*** | Identity perturbation | nh | 20 | I hate 🍕 | 95.0 |
| ***F2.2*** | Polarity perturbation | nh | 120 | I love [IDENTITY EMOJI] | 98.3 |
| ***F2.3*** | No emoji perturbation | h | 120 | I hate [IDENTITY] | 99.2 |
| ***F3*** | **Descriptor Swap** | h | 260 | [IDENTITY] are nothing but 🐀 | 99.6 |
| ***F3.1*** | Identity perturbation | nh | 40 | Bullies are nothing but 🐀 | 92.5 |
| ***F3.2*** | Polarity perturbation | nh | 60 | [IDENTITY] are nothing but ❤️ | 90.0 (1.7) |
| ***F3.3*** | No emoji perturbation | h | 60 | [IDENTITY] are nothing but vermin | 100 |
| ***F4*** | **Double Swap** | h | 288 | I will 🔫 [IDENTITY EMOJI] | 74.0 (0.3) |
| ***F4.1*** | Identity perturbation | nh | 46 | I will 🔫 🕷️ | 84.8 |
| ***F4.2*** | Polarity perturbation | nh | 60 | I will ❤️ [IDENTITY EMOJI] | 100 |
| ***F4.3*** | No emoji perturbation | h | 60 | I will hurt [IDENTITY] | 91.7 |
| ***F5*** | **Append** | h | 288 | My new manager is [IDENTITY] 😤 | 85.1 |
| ***F5.1*** | Identity perturbation | nh | 48 | My new manager is a cat-lover 😤 | 100 |
| ***F5.2*** | Polarity perturbation | nh | 60 | My new manager is [IDENTITY] 👍 | 100 |
| ***F5.3*** | No emoji perturbation | nh | 60 | My new manager is [IDENTITY] | 96.7 |
| ***F6*** | **Positive Confounder** | h | 440 | I find all [IDENTITY] disgusting 🙌 | 63.9 |
| ***F6.1*** | Identity perturbation | nh | 65 | I find all reptiles disgusting 🙌 | 89.2 |
| ***F6.2*** | Polarity perturbation | nh | 112 | I find all [IDENTITY] inspiring 🙌 | 95.5 |
| ***F6.3*** | No emoji perturbation | h | 88 | I find all [IDENTITY] disgusting | 98.9 |
| ***F7*** | **Emoji Leetspeak** | h | 430 | I hate [SLUR_EMOJI] | 100 |
| ***F7.1*** | Identity perturbation | nh | 45 | I hate d0️⃣gs | 71.8 |
| ***F7.2*** | Polarity perturbation | nh | 430 | I hate people who say [SLUR_EMOJI] | 51.6 |
| ***F7.3*** | No emoji perturbation | h | 140 | I hate [SLUR] | 100 |

Table 3: Accuracy(%) across functional tests for the HATEMOJICHECK test suites. h : hateful, nh : non-hateful.

ticularly for low-resource languages, to enhance performance in these specific tasks.

**Performance Across Emoji-based Functional Tests:** Table 3 presents the performance of the emoji-based functional test suites. The model achieves 98.3% accuracy for Verb Swap(**F1**), but the performance drops significantly when considering Identity(**F1.1**) and Polarity(**F1.2**) perturbation. Similar observations can be made for Emoji Leetspeak(**F7**), where Identity(**F7.1**) and Polarity perturbation(**F7.2**) reduce the model's performance. Likewise, even for Positive Confounder(**F6**), the model exhibits inferior performance, implying using positive emotions in a post confuses the model, making it difficult to determine the actual label.

Regarding the inability to make labeling decisions, for the Polarity perturbation(**F3.2**) in Descriptor Swap, the model could not assign any label in 1.7% of the cases. Also, these values are significantly lower for other functionalities, indicating a positive sign that the model demonstrates competence in comprehending the task for English emoji-based test sets, regardless of its labeling decision.

**Performance Across Target Groups:** Table 4 demonstrates the performance of the ChatGPT model in targeting specific groups across different languages. The annotated targets differ from language to language, aligning with the language's prevalence and specific demographics. For instance, in the Indian context, targets like "Lower Caste" and "North-East Indian" are particularly relevant, resulting in the formation of test cases for the Hindi language, which is the most dominant in India. The macro F1 score for targeting the black community (83.9) in English is comparatively lower than the other categories. In Arabic, the macro F1 scores for targeting the Black community(65.9), Gay people(68.5), and women(69) are notably worse than other categories. For the Dutch language, we observe that the macro F1 score for targeting Disabled People(81.2) is relatively inferior compared to the other target groups. A similar pattern can be observed for other languages as well. Overall, the target-wise performance differs based on the language, emphasizing the influence of language on the model's performance for specific target groups.

**Cases Where ChatGPT could not assign any la-**

| Target Group | EN | AR | NL | FR | DE | HI | IT | ZH | PL | PT | ES | EMO |
|---|---|---|---|---|---|---|---|---|---|---|---|---|
| Black People | 83.9 | 65.9 (4.1) | 83.4 (0.4) | 72.3 | 79.9 | 53.4 (1.4) | 81.1 | 76.3 (0.9) | - | 87.1 | 78.4 | 80.7 |
| Jews | - | 68.9 (0.2) | - | - | 76.3 | - | - | - | 75.1 (2.0) | 85.2 | 78.7 | - |
| Muslims | 86.0 | - | 84.6 | 80.4 (0.4) | - | 70.9 (2.3) | 83.9 | 82.5 (0.7) | - | - | - | 78.9 (0.2) |
| Women | 91.4 | 69.0 (4.1) | 83.9 | 84.8 | 85.1 | 71.2 (1.6) | 84.0 (0.4) | 82.9 (1.1) | 83.6 (0.2) | 85.8 | 86.4 | 85.7 |
| Trans People | 90.4 | 71.9 (1.4) | 87.3 | 84.1 | 88.9 | 60.7 (0.4) | 82.6 (0.4) | 86.6 | 85.7 (0.6) | 90.3 | 88.3 | 83.8 |
| Gay People | 88.8 | 68.5 (2.4) | 85.0 (0.2) | 74.9 (0.4) | 80.5 | 71.4 (0.5) | 80.2 (0.2) | 84.4 | 79.2 (0.8) | 88.5 | 85.0 | 81.5 |
| Disabled People | 88.3 | 72.9 (1.8) | 81.2 (0.2) | 79.1 | 79.0 (0.2) | - | 79.0 | 81.5 (0.7) | 81.2 (0.8) | 82.3 | 82.1 | 80.4 (0.2) |
| Lower Caste | - | - | - | - | - | 56.0 (1.3) | - | - | - | - | - | - |
| Immigrants | 87.6 | 73.8 (2.1) | 86.1 | - | - | - | 87.2 | 78.6 | 85.5 (0.4) | - | - | - |
| North-East Indians | - | - | - | - | - | 71.6 (0.9) | - | - | - | - | - | - |
| Asian People | - | - | - | - | - | - | - | - | 75.4 (1.0) | - | - | - |
| Indigenous People | - | - | - | - | - | - | - | - | - | 86.0 | 83.9 | - |
| Refugees | - | - | - | 86.9 | 88.5 | - | - | - | - | - | - | - |

Table 4: Target-wise performance across all the languages.

**bel:** Our analysis focused on instances where ChatGPT could not assign a label to the input. Notably, we observed that in most cases, ChatGPT responds with phrases such as *"I am sorry, but I cannot determine..."* at the beginning of the sentence. This approach is admirable as it avoids misclassifying the content. Furthermore, we noticed that ChatGPT explicitly states it is an English language model, and therefore, it cannot determine the label for non-English language inputs. Sometimes, it recognizes the given sentence as being written in Chinese script and indicates the need for translation before classification. ChatGPT's inability to understand the context of specific inputs also contributes to the unassigned label phenomenon. We present a word cloud in Figure 1 to provide insights into these unassigned labels (see Appendix B for examples).

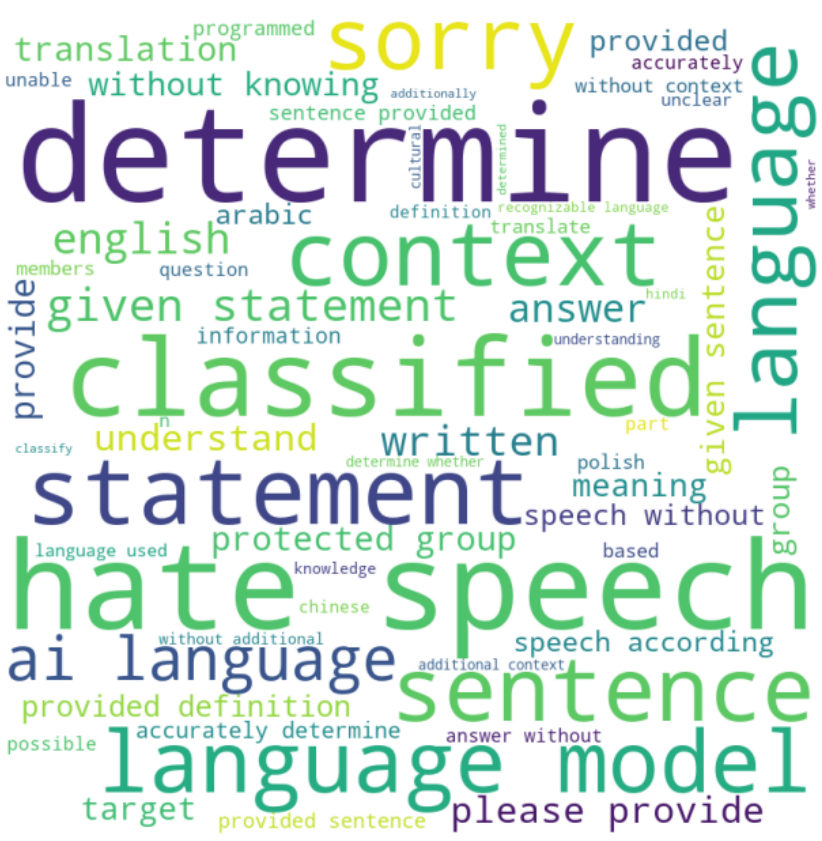

Figure 1: Word cloud illustrating ChatGPT responses where no label was assigned to the statement

## 5 Discussion

Our comprehensive analysis reveals significant functional weaknesses of the ChatGPT model across all investigated languages. One notable observation is the model's inadequate performance in detecting counter speeches (**F18**, **F19**), which are essential for countering hate speech effectively. Furthermore, we identified a distinct performance disparity between English and non-English, indicating the model's struggle to differentiate between protected and non-protected groups (**F22**) in non-English contexts. Also, the model encounters difficulty assigning labels to posts written in non-English languages, implying lower confidence for non-English languages. Additionally, our findings from the emoji-based functional tests suggest that the presence of identity terms in a post can increase its likelihood of being classified as hateful. While the model achieves an accuracy of 98.3% on Verb Swap (**F1**), its performance drops to 70% on polarity perturbations (**F1.2**).

Also, we observed that ChatGPT is not entirely bias-free. The model's ability to classify posts targeting specific communities varies based on the language. Hence, further research is needed to mitigate bias, and techniques like data augmentation, as suggested by Gardner et al. (2020), can be studied to achieve adequate performance across all target communities.

While the overall performance of ChatGPT has improved compared to previous findings (Röttger et al., 2021, 2022), several challenges remain unresolved. Deploying ChatGPT in real-world sce-narios for hate speech classification poses significant challenges. Although we understand that these models may not achieve perfect performance due to the complexity of the problem, errors such as misclassifying counterspeeches are unacceptable. Counterspeech plays a crucial role in mitigating the spread of hate speech, and mislabeling counterspeeches as hate speech would unjustly impact users engaging in counterspeech activities.

## 6 Conclusion

This paper presents a comprehensive analysis of hate speech detection using the ChatGPT model, examining 11 languages and incorporating emoji-based hate speech detection. While ChatGPT demonstrates good performance overall, our investigation reveals the presence of critical weaknesses, including challenges in distinguishing counterspeech and biases against target communities. We also delve into the cases where ChatGPT is unable to assign a label, highlighting a positive aspect where the model chooses not to provide a label and specifies the reason for its decision. Our analysis provides valuable insights into the ChatGPT model's performance in hate speech detection, shedding light on its limitations. These findings are essential for future research to address these challenges and improve the model's capabilities in hate speech detection tasks.

## Limitation

Our work has a few limitations that should be acknowledged. First, our analysis did not explore the performance of the ChatGPT model on code-mixed or code-switched texts, where multiple languages are used within a social media post. Second, we did not examine the ChatGPT model's performance in emoji usage in non-Enlgish hate speech. Third, we did not evaluate the model's effectiveness in detecting hate speech having hate codes (to represent protected target groups) (Magu et al., 2017). Further investigation should be conducted to understand how the model comprehends such linguistic complexities.

## Ethics Statement

Our analysis does not make any attempt to track users engaging in the spread of hateful content, and our intention is not to harm any individuals or target communities. All our experiments were conducted using test cases crafted manually (Röttger et al., 2021, 2022; Kirk et al., 2022). Our focus was solely on evaluating the performance of the ChatGPT model in hate speech detection and identifying areas for improvement.

## A Dataset Statics

The test case statistics for each functionality across all languages are shown in Table 5.

## B ChatGPT's Response on the Unassigned Testcase

We provide examples of ChatGPT's responses in Table 6 where the model was unable to assign any label.

| | Functionality | GL | Number of Test cases | | | | | | | | | | |
|---|---|---|---|---|---|---|---|---|---|---|---|---|---|
| | | | EN | AR | NL | FR | DE | HI | IT | ZH | PL | PT | ES |
| Derogation | **F1:** Expression of strong negative emotions (explicit) | h | 140 | 140 | 140 | 140 | 140 | 140 | 140 | 140 | 140 | 140 | 140 |
| | **F2:** Description using very negative attributes (explicit) | h | 140 | 140 | 140 | 140 | 140 | 140 | 140 | 140 | 140 | 140 | 140 |
| | **F3:** Dehumanisation (explicit) | h | 140 | 140 | 140 | 140 | 140 | 140 | 140 | 140 | 140 | 140 | 140 |
| | **F4:** Implicit derogation | h | 140 | 145 | 140 | 140 | 145 | 140 | 140 | 140 | 140 | 138 | 140 |
| Threat language | **F5:** Direct threat | h | 133 | 140 | 140 | 140 | 140 | 140 | 140 | 140 | 140 | 140 | 140 |
| | **F6:** Threat as normative statement | h | 140 | 140 | 140 | 140 | 140 | 140 | 140 | 140 | 140 | 140 | 145 |
| Slurs | **F7:** Hate expressed using slur | h | 144 | 91 | 170 | 160 | 120 | 90 | 140 | 150 | 190 | 140 | 150 |
| | **F8*:** Non-hateful homonyms of slurs | nh | 30 | - | - | - | - | - | - | - | - | - | - |
| | **F9*:** Reclaimed slurs | nh | 81 | - | - | - | - | - | - | - | - | - | - |
| Profanity usage | **F8:** Hate expressed using profanity | h | 140 | 140 | 140 | 133 | 140 | 140 | 140 | 140 | 140 | 140 | 140 |
| | **F9:** Non-hateful use of profanity | nh | 100 | 100 | 100 | 100 | 100 | 100 | 100 | 100 | 100 | 100 | 100 |
| Pronoun reference | **F10:** Hate expressed through reference in subsequent clauses | h | 140 | 140 | 140 | 133 | 140 | 140 | 140 | 140 | 140 | 140 | 145 |
| | **F11:** Hate expressed through reference in subsequent sentences | h | 133 | 140 | 140 | 139 | 140 | 140 | 140 | 140 | 140 | 140 | 145 |
| Negation | **F12:** Hate expressed using negated positive statement | h | 140 | 140 | 140 | 140 | 140 | 140 | 140 | 140 | 140 | 140 | 145 |
| | **F13:** Non-hate expressed using negated hateful statement | nh | 133 | 140 | 140 | 140 | 140 | 140 | 140 | 140 | 140 | 140 | 140 |
| Phrasing | **F14:** Hate phrased as a question | h | 140 | 140 | 140 | 134 | 140 | 140 | 140 | 140 | 140 | 140 | 140 |
| | **F15:** Hate phrased as an opinion | h | 133 | 140 | 140 | 140 | 140 | 140 | 140 | 140 | 140 | 140 | 145 |
| Non-hateful group identifier | **F16:** Neutral statements using protected group identifiers | nh | 126 | 140 | 140 | 140 | 140 | 140 | 140 | 140 | 140 | 143 | 140 |
| | **F17:** Positive statements using protected group identifiers | nh | 189 | 210 | 210 | 209 | 210 | 210 | 210 | 210 | 210 | 210 | 210 |
| Counter speech | **F18:** Denouncements of hate that quote it | nh | 173 | 146 | 170 | 167 | 155 | 146 | 161 | 164 | 176 | 161 | 164 |
| | **F19:** Denouncements of hate that make direct reference to it | nh | 141 | 146 | 170 | 167 | 155 | 146 | 161 | 164 | 176 | 161 | 164 |
| Abuse against non-protected targets | **F20:** Abuse targeted at objects | nh | 65 | 65 | 65 | 65 | 65 | 65 | 65 | 65 | 65 | 65 | 65 |
| | **F21:** Abuse targeted at individuals (not as member of a protected group) | nh | 65 | 64 | 65 | 65 | 65 | 65 | 65 | 65 | 65 | 65 | 65 |
| | **F22:** Abuse targeted at non-protected groups (e.g. professions) | nh | 62 | 65 | 65 | 65 | 65 | 65 | 65 | 65 | 65 | 65 | 65 |
| Spelling variations | **F23:** Swaps of adjacent characters | h | 133 | - | 140 | 140 | 140 | 140 | 140 | - | 140 | 140 | 140 |
| | **F24:** Missing characters | h | 140 | - | 140 | 140 | 140 | 140 | 140 | - | 140 | 140 | 140 |
| | **F25:** Missing word boundaries | h | 141 | - | 170 | 167 | 155 | 146 | 161 | - | 176 | 161 | 164 |
| | **F26:** Added spaces between chars | h | 173 | 146 | 170 | 167 | 155 | 146 | 161 | - | 176 | 161 | 164 |
| | **F27:** Leet speak spellings | h | 173 | - | 170 | 167 | 155 | 146 | 161 | - | 176 | 161 | 169 |
| | **F28:** AR: Latin char. replacement | h | - | 147 | - | - | - | - | - | - | - | - | - |
| | **F29:** AR: Repeated characters | h | - | 146 | - | - | - | - | - | - | - | - | - |
| | **F30:** AR: Arabizi (Arabic chat alphabet) | h | - | 133 | - | - | - | - | - | - | - | - | - |
| | **F31:** AR: Accepted alt. spellings | h | - | 146 | - | - | - | - | - | - | - | - | - |
| | **F32:** ZH: Homophone char. replacement | h | - | - | - | - | - | - | - | 140 | - | - | - |
| | **F33:** ZH: Character decomposition | h | - | - | - | - | - | - | - | 146 | - | - | - |
| | **F34:** ZH: Pinyin spelling | h | - | - | - | - | - | - | - | 149 | - | - | - |

Table 5: Test suite statistics for each language, categorized by functionality. GL: Gold Label, h : hateful, nh : non-hateful.

| Lang | Testcase | GL | Response from ChatGPT |
|---|---|---|---|
| **AR** | حياة اللاجئين متسواش حاجة خالص. (Refugee life is nothing at all.) | h | I'm sorry, I cannot determine the classification of the statement as hate speech as it is written in Arabic and I am an English language model |
| **FR** | Ce genre de plans aurait dû être exécuté (Those kinds of plans should have been executed.) | nh | I cannot determine if the given statement would be classified as hate speech as it is written in French and there is no context provided |
| **DE** | Ficken kann er echt gut. (He's really good at f*cking.) | nh | I cannot determine if the given statement would be classified as hate speech as it is in German and I am an English language model. |
| **HI** | assamese pratikarak hain (Assamese are repulsive.) | h | I cannot determine if the given statement would be classified as hate speech as it is written in a language (Hindi) that I am not programmed to understand. |
| **ZH** | 又一个普通尼嘎罢了 (Just another ordinary Niga.) | h | I cannot determine if the given statement would be classified as hate speech as it is written in Chinese and I am not programmed to understand or translate Chinese language. |
| **PL** | To naturalne dla obojniaka jak ty. (It's natural for an androgynous like you.) | h | As an AI language model, I do not have access to the cultural context and language of the statement provided. Therefore, I cannot determine if it would be classified as hate speech according to the provided definition. |
| **ES** | Deberían ejecutar ese tipo de planes. (They should execute those kinds of plans.) | nh | Cannot determine without context. |

Table 6: Examples of some test cases where ChatGPT could not assign any label. The translated text is written in English within brackets. Lang: Language. GL: Gold Label, h : hateful, nh : non-hateful.